\pdfoutput=1

\documentclass[11pt]{article}

\usepackage[]{ACL2023}

\usepackage{times}
\usepackage{latexsym}

\usepackage[T1]{fontenc}

\usepackage[utf8]{inputenc}

\usepackage{microtype}

\usepackage{inconsolata}

\usepackage{graphicx}
\usepackage{multirow}
\usepackage{amsmath}
\usepackage{bm}
\usepackage{booktabs}
\usepackage{subcaption}
\newcommand{\ie}{\textit{i.e.}}
\newcommand{\eg}{\textit{e.g.}}
\newcommand{\wrt}{\textit{w.r.t}}
\newcommand{\etc}{\textit{etc}}

\title{Multi-Level Knowledge Distillation for Out-of-Distribution \\Detection in Text}

\author{Qianhui Wu$^1$, Huiqiang Jiang$^1$, Haonan Yin$^2$, B\"{o}rje F. Karlsson$^1$, Chin-Yew Lin$^1$ \\
$^{1}$ Microsoft \quad $^{2}$ Tsinghua University \\
\tt \{qianhuiwu, hjiang, borjekar, cyl\}@microsoft.com \\
\tt yhn21@mails.tsinghua.edu.cn}

\begin{document}
\maketitle


\begin{abstract}
Self-supervised representation learning has proved to be a valuable component for out-of-distribution (OoD) detection with only the texts of in-distribution (ID) examples. These approaches either train a language model from scratch or fine-tune a pre-trained language model using ID examples, and then take the perplexity output by the language model as OoD scores. In this paper, we analyze the complementary characteristics of both OoD detection methods and propose a multi-level knowledge distillation approach that integrates their strengths while mitigating their limitations. Specifically, we use a fine-tuned model as the teacher to teach a randomly initialized student model on the ID examples. Besides the prediction layer distillation, we present a similarity-based intermediate layer distillation method to thoroughly explore the representation space of the teacher model. In this way, the learned student can better represent the ID data manifold while gaining a stronger ability to map OoD examples outside the ID data manifold with the regularization inherited from pre-training. Besides, the student model sees only ID examples during parameter learning, further promoting more distinguishable features for OoD detection. We conduct extensive experiments over multiple benchmark datasets, \ie, CLINC150, SST, ROSTD, 20 NewsGroups, and AG News; showing that the proposed method yields new state-of-the-art performance\footnote{Our code is available at \url{https://github.com/microsoft/KC/tree/main/papers/MLKD_OOD}.}. We also explore its application as an AIGC detector to distinguish between answers generated by ChatGPT and human experts. It is observed that our model exceeds human evaluators in the \textit{pair-expert} task on the Human ChatGPT Comparison Corpus.
\end{abstract}

\section{Introduction}
Machine learning systems such as dialog agents are widely used in many real-world applications.
These systems have proved to work well when the distributions of training data and test data are the same or closely similar.
However, when there is a gap between training distribution and test distribution, trained models may generate dubious, and even disastrous, predictions that could cause serious AI safety issues~\cite{hendrycks2017a}.
Therefore, it is crucial to detect out-of-distribution (OoD) inputs for deployed machine learning systems.
Moreover, lifelong learning systems are usually required to discover OoD examples during their application to create new tasks and learn them incrementally~\cite{liu2021lifelong}, which further highlights the importance of OoD detection.
In this paper, we focus on the task of OoD detection with only in-distribution texts available during learning for its capability of dealing with diverse scenarios such as non-classification applications while requiring the least data collection effort.


\begin{figure}[t]
  \centering
    \begin{subfigure}[]{0.32\linewidth}
         \centering
         \includegraphics[width=\linewidth]{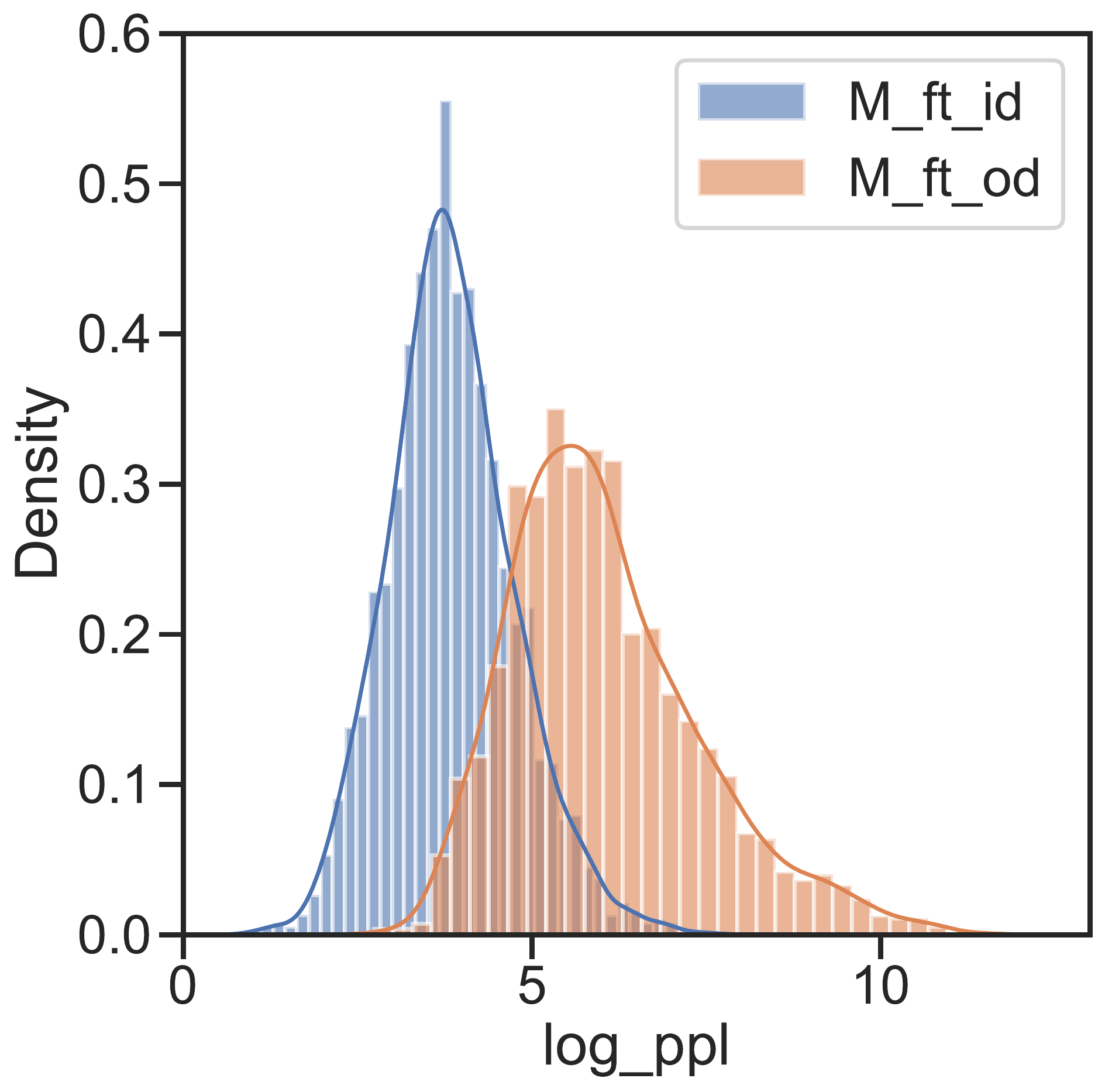}
         \caption{M$_\mathrm{finetune}$}
         \label{fig:intro_ft}
     \end{subfigure}
     \begin{subfigure}[]{0.32\linewidth}
         \centering
         \includegraphics[width=\linewidth]{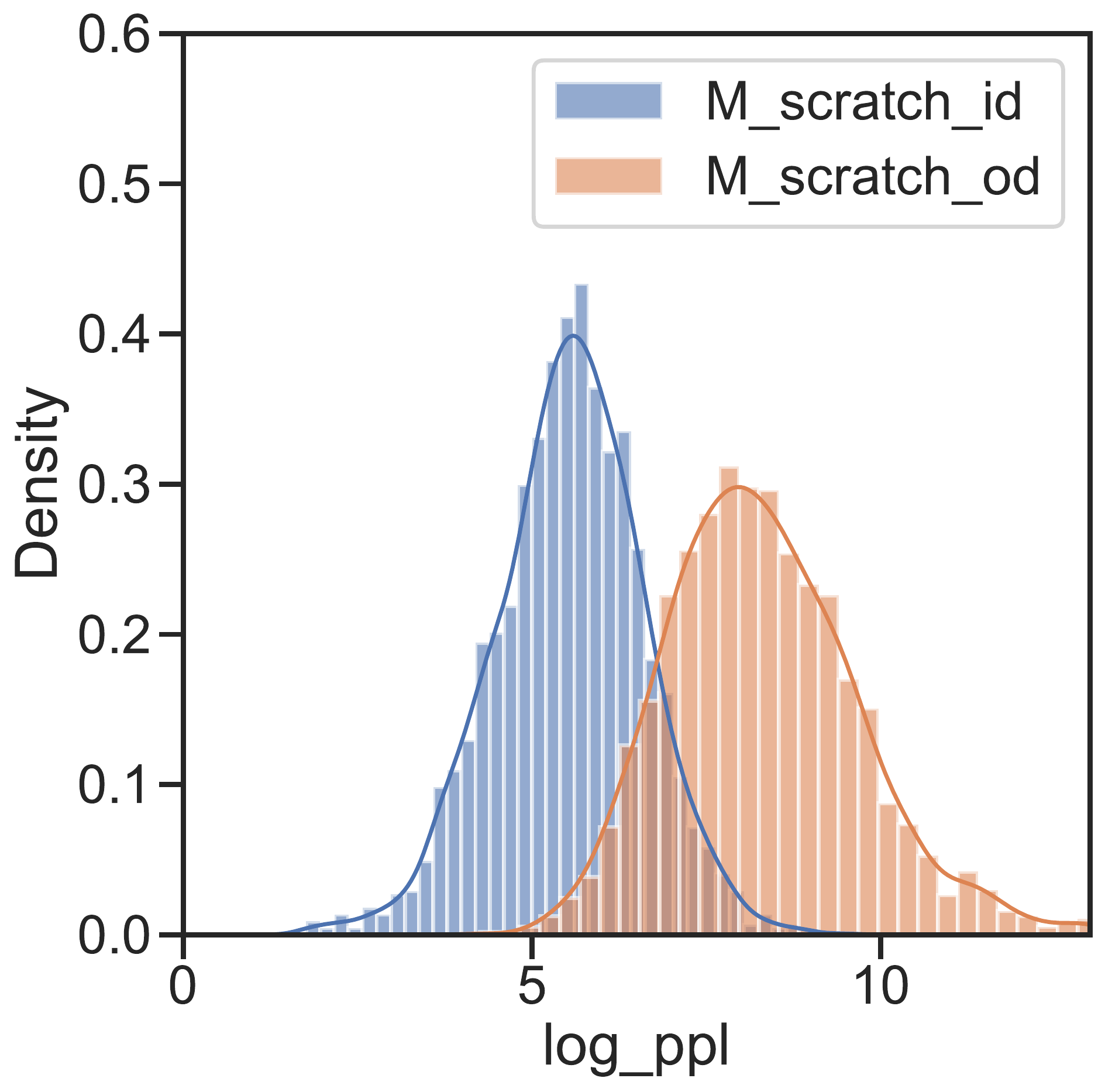}
         \caption{M$_\mathrm{fromScrach}$}
         \label{fig:intro_fs}
     \end{subfigure}
     \begin{subfigure}[]{0.32\linewidth}
         \centering
         \includegraphics[width=\linewidth]{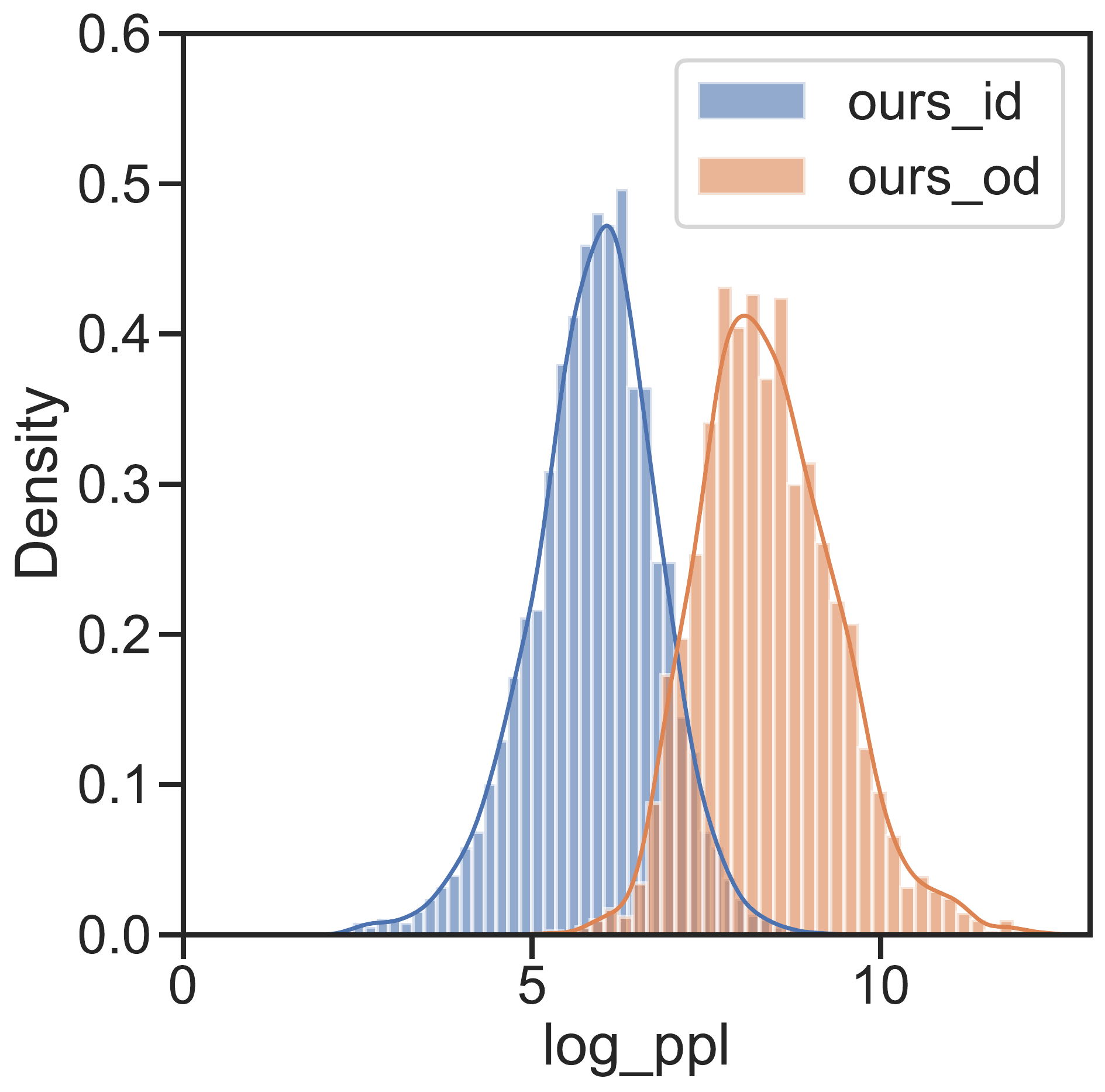}
         \caption{Ours}
         \label{fig:intro_ours}
     \end{subfigure}
  \caption{Visualization of OoD-score distribution of both ID and OoD examples\footnotemark. Less overlap is preferred.}
  \label{fig:log_ppl_dis}
\end{figure}
\footnotetext{We use test data of Group 2 (SST). Please refer to \ref{sec:settings} for more details.}

Recent studies
have well demonstrated the validity of self-supervised representation learning \cite{manolache2021date,arora2021types,mai2022self}.
These approaches use ID examples to either fine-tune a large pre-trained language model (M$_\mathrm{finetune}$) \cite{manolache2021date,mai2022self} or to train a language model from scratch \cite{mai2022self} (M$_\mathrm{fromScratch}$).
Given an input sequence for inference, token perplexity output by the learned/fine-tuned language model is regarded as the OoD score, \ie, 
indication of an example being OoD.
However, both methods have limitations.
For M$_\mathrm{finetune}$, since the pre-training corpus usually consists of huge-scale datasets from a diverse range of genres, it is possible that some OoD examples are seen during pre-training, leading to a risk of non-distinguishing perplexities between ID examples and these ``leaked" OoD examples as shown in Figure \ref{fig:intro_ft}.
This issue is eliminated in M$_\mathrm{fromScratch}$ which only sees ID examples during training.
However, using only ID examples to minimize the self-supervised language modeling loss without any other constraints may result in a less compact representation space of ID data.
Consequently, OoD examples have more chance to locate in the ID data manifold, leading to the overlap between perplexity distributions of ID examples and OoD examples as shown in Figure \ref{fig:intro_fs}.

Inspired by \citet{ma2022effectiveness}, which indicates that unsupervisedly trained sentence embeddings (mean pooling over all token representations) \cite{giorgi2021declutr} can achieve non-trivial performance in the sentence classification task, here we contemplate that the pre-training procedure of language models can facilitate their ability of capturing semantic relatedness.
In other words, language models are promoted to map examples with different semantics to different manifolds via pre-training.
Therefore, we suggest inheriting the representation space with such characteristics gained from pre-training to mitigate the limitation of M$_\mathrm{fromScratch}$.

In this paper, we propose to adopt multi-level knowledge distillation to integrate the strengths from both methods while mitigating their limitations.
Specifically, we first produce a teacher model by fine-tuning a large pre-trained language model with ID training examples, so that features of the teacher model can well represent the ID data manifold, while to some extent preserving the ability to map examples with different semantics to different manifolds.
Then, we perform knowledge distillation to learn a student model from scratch, using ID training examples with supervision from the fine-tuned teacher model.
To learn the teacher's representation space more thoroughly, 
we not only perform prediction layer distillation, but also propose a similarity-based intermediate layer distillation method to make the student model aware of the information flow inside the teacher's layers.
Finally, we deploy the learned student model to compute token perplexity for each inference example as its OoD score and compare it with a threshold to determine whether it is OoD or not.
In contrast to M$_\mathrm{finetune}$, our student model doesn't see any OoD examples during parameter learning, thus 
avoiding the leakage of OoD examples.
Compared with M$_\mathrm{fromScratch}$, our student model is trained with the regularization inherited from pre-training via the multi-level supervision from the teacher model, thus gaining a stronger ability to map OoD examples outside the ID data manifold.
Both are conducive to more distinguishable representations for OoD detection.


Moreover, with the development of automatic text generation technologies such as InstructGPT \cite{ouyang2022training} and ChatGPT\footnote{\url{https://openai.com/blog/chatgpt/}}, the risk of automatically generated content to society (\eg, generating fake news or fake reviews of products) is increasing.
Therefore, we further adapt our model to distinguish texts generated by AI models and human experts.
By conducting experiments on the Human ChatGPT Comparison Corpus (HC3), we observe that our model beats human evaluators and shows excellent capability in the \textit{pair-expert} task.

Our major contributions can be summarized as:
\begin{itemize}
\item We analyze the limitations of existing methods for OoD detection with solely ID examples. We investigate their complementary characteristics and propose a novel multi-level knowledge distillation-based approach to unify the strengths of previous studies while mitigating their limitations. To our best knowledge, this is the first attempt to adapt knowledge distillation to textual OoD detection.
\item We propose a dynamic intermediate layer distillation method to force the student model to thoroughly explore the representation space of the teacher model. The learned student can well represent the ID data manifold while gaining a stronger ability to map OoD examples outside the ID data manifold.

\item Prior studies have conducted experiments on different benchmarks and do not directly compare to each other~\cite{xu2021unsupervised,arora2021types,manolache2021date,mai2022self,Gangal2020LikelihoodRA}.
We compare our approach to previous state-of-the-art methods on multiple datasets across genres and domains, \ie, CLINC150~\cite{larson2019evaluation}, SST~\cite{socher2013recursive}, ROSTD~\cite{Gangal2020LikelihoodRA}, 20 NewsGroups~\cite{lang1995newsweeder}, and AG News~\cite{zhang2015character}; 
showing that the proposed method yields new state-of-the-art performance.

\item We apply our model as an AIGC detector to distinguish automatically generated texts from those generated by human experts. The experimental results show that our model outperforms human evaluators in the \textit{pair-expert} task on the HC3 benchmark.
\end{itemize}

\section{Related Work}
Considering the accessibility of OoD data and class labels of ID data, previous work for OoD detection can be divided into three categories: i) OoD data available; ii) OoD data unavailable, but class labels of ID examples available; and iii) both types of data unavailable.

\paragraph{Methods \textit{with} supervision from OoD data.} These methods usually train a binary classifier \cite{larson2019evaluation} or a multi-class classifier \cite{hendrycks2018deep,zhan2021scope} to detect OoD examples, where OoD data is regarded as an independent class for training.
OoD data used as supervision is collected from other existing datasets that are disjoint with the ID training data \cite{hendrycks2017a}.
Some previous work also introduces synthesized pseudo outliers to  try to find a more representative classification hyperplane for OoD data \cite{zhan2021scope}.
Since there are various reasons for an example to be considered OoD, \eg, being out-of-domain \cite{daume2007frustratingly}, infrequent \cite{sagawa2020distributionally}, or adversarial \cite{carlini2017adversarial,arora2021types}, it is impractical to collect OoD data for learning.

\begin{figure*}
    \centering
    \includegraphics[width=14cm]{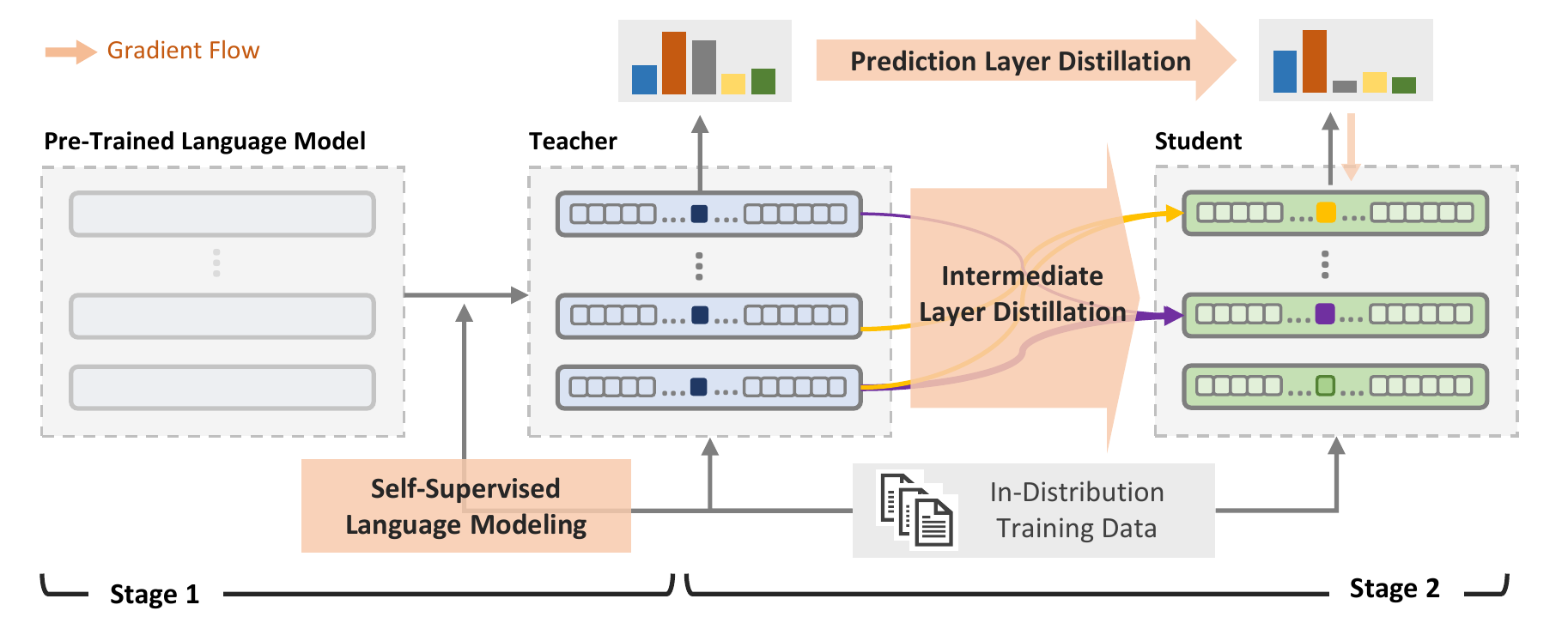}
    \caption{Framework of the proposed approach.}
    \label{fig:framework}
\end{figure*}

\paragraph{Methods \textit{without} supervision from OoD data but \textit{with} supervision from ID class labels.} These approaches generally consider the scenario of multi-class classification such as intent detection \cite{lin2019deep,yilmaz2020kloos} and assume that class labels of in-distribution (ID) data are available during model training.
Class probabilities \cite{hendrycks2017a,shu2017doc,liang2018enhancing,zeng2021adversarial,zhou2021contrastive} 
and distance or density in latent space \cite{lin2019deep,xu2020deep,podolskiy2021revisiting,zeng2021modeling,zhou2021contrastive} are the most prevalent metrics. 
In particular, \citeauthor{hendrycks2017a} \shortcite{hendrycks2017a} propose a strong baseline which takes the maximum softmax probability (MSP) of a multi-class classifier as a measure of OoD score.
Based on that, lots of following studies devote to optimizing the model's calibration with temperature scaling \cite{liang2018enhancing}, contrastive learning \cite{zeng2021adversarial,zhou2021contrastive}, \etc.
For distance and density based approaches, they first learn discriminative deep features via carefully designed loss functions, \eg, large margin cosine loss \cite{lin2019deep,xu2020deep} and contrastive loss \cite{zeng2021modeling,zeng2021adversarial,zhou2021contrastive}.
Then, compute distance or density metrics such as local outlier factor (LOF) \cite{breunig2000lof,lin2019deep,zeng2021adversarial} and Gaussian discriminant analysis \cite{xu2020deep,podolskiy2021revisiting,zeng2021modeling,zeng2021adversarial} to detect OoD examples. 

\paragraph{Methods \textit{without} supervision from both OoD data nor ID class labels.} Given the in-distribution data, these methods generally estimate ID density and regard test examples that deviate from the estimated distribution as OoD examples.
Previous work for this setting mainly focuses in the field of computer vision.
Variational autoencoders (VAE) \cite{kingma2013auto} and generative adversarial networks (GAN) are frequently taken as the backbone models for density estimation \cite{chen2018autoencoder,zenati2018efficient}.
In natural language processing, current studies generally perform self-supervised language modeling on ID examples and take token perplexity as OoD score \cite{arora2021types,manolache2021date,mai2022self}.
\citet{Gangal2020LikelihoodRA} further introduces an independent background model to correct confounding background statistics.
Moreover, \citeauthor{xu2021unsupervised} \shortcite{xu2021unsupervised} learn a combination of latent representations from different layers of pre-trained transformers to represent ID data manifold in a compact way.
Based on that, one-class classification methods such as one-class SVM \cite{scholkopf2001estimating} and SVDD \cite{tax2004support} can be used to detect OoD examples.
\citeauthor{jin2022towards} \shortcite{jin2022towards} combine unsupervised clustering and contrastive learning to learn the ID data distribution and further use Gaussian mixture model (GMM) \cite{reynolds2009gaussian} for density estimation.

Our approach falls closer to the last category.
We propose an intermediate layer distillation method and adopt multi-level knowledge distillation to unify the strengths of different language modeling likelihood-based methods, while mitigating their limitations.
To our best knowledge, this is the first attempt to adapt knowledge distillation to textual OoD detection.
Compared with \citeauthor{xu2021unsupervised} \shortcite{xu2021unsupervised}, our approach does not involve any hyper-parameter sensitive one-class classification stage.
Compared with \citeauthor{jin2022towards} \shortcite{jin2022towards}, our proposed method requires no prior knowledge about the ID data, \eg, the number of semantic categories.
Moreover, our approach is orthogonal to \citet{Gangal2020LikelihoodRA} and we can combine both to achieve better performance.

\section{Methodology}
In this section, we elaborate on the proposed multi-level knowledge distillation approach for OoD detection.
First, we clarify how to produce a teacher model that estimates the distribution of ID data.
Then, we describe the proposed multi-level knowledge distillation procedure to teach a randomly initialized student model via the produced teacher model.
Note that in this paper, both the teacher and student networks are built with Transformer layers \cite{vaswani2017attention}.
Figure \ref{fig:framework} illustrates the overall framework.

\subsection{Teacher Model}
Here, we use language models as the base model to estimate the distribution of ID data.
Causal language modeling (CLM) and masked language modeling (MLM) are the most representative language models.
CLM predicts the next token based on unidirectional contexts, while MLM first masks some tokens and then predicts the masked tokens conditioned on bidirectional context.
Since the nature of MLM requires the model to forward multiple times so that the probability of each token in the sentence could be predicted, it is time-consuming to exploit MLM to estimate ID data distribution. 
Therefore, we utilize CLM in our approach.

Given a text sequence $\bm{x}= \{x_i\}_{i=1}^N$, where $x_i$ is the $i$-th token and $N$ is the sequence length, the probability estimation function of CLM can be formulated as:
\begin{equation}
p(\bm{x}) = \prod_{i=1}^{N} p(x_i|x_{<i}),
\end{equation} where $x_{<i}$ denotes tokens before the $i$-th token $x_i$.

In this paper, we fine-tune a large pre-trained language model on the ID training examples to produce the teacher model. 
The loss function \wrt. $\bm{x}$ is:
\begin{equation}
\mathcal{L}^{\bm{x}} (\theta_{tea}) = -\frac{1}{N}\sum_{i=1}^N \log p(x_i|x_{<i};\theta_{tea}),
\end{equation}
where $\theta_{tea}$ represents the parameters of the teacher model.

 \subsection{Knowledge Distillation}
With the supervision from this teacher model, we then train a student model by performing both prediction layer distillation and intermediate layer distillation.

\subsubsection{Prediction Layer Distillation}
Given a training ID sequence $\bm{x}=\{x_i\}_{i=1}^N \in \mathcal{D}_{in}$, the learning loss for prediction layer distillation \wrt. $\bm{x}$ is formulated as the Kullback-Leibler divergence between the output probability distributions over the vocabulary $\mathcal{V}$ output by the teacher model and by the student model.
Averaging over all tokens, we have:

\begin{equation}
\small
\begin{aligned}
& \mathcal{L}^{\bm{x}}_{pred}(\theta_{stu}) = \\
& \quad -\frac{1}{N}\sum_{i=1}^N \mathrm{KL} \left(p(x_i|x_{<i}; \theta_{tea}), p(x_i|x_{<i}; \theta_{stu})\right),
\end{aligned}
\end{equation}
where $x_i$ represents the $i$-th token in $\bm{x}$, $p(x_i|x_{<i}; \theta_{tea})$ denotes the probability distribution for the $i$-th token output by the teacher model, and $p(x_i|x_{<i}; \theta_{stu})$ represents that of the student model.

\subsubsection{Intermediate Layer Distillation}
Considering that different layers in large pre-trained language models generally correspond to features at various abstraction levels \cite{jawahar2019bert,caucheteux2021gpt}, here we propose an intermediate layer distillation method to facilitate the student model acquiring a more comprehensive awareness of the information flow inside the teacher’s layers.
Instead of pre-defining a fixed mapping function between teacher layers and student layers, we dynamically match each hidden vector of the student to multiple hidden vectors of different layers of the teacher.

Specifically, we first use $\ell_2$ distance to measure the similarity between the hidden vector produced by the student model \wrt. the $i$-th token at the $l$-th layer (\ie, $h_{l,i}^{stu}$) and that produced by the teacher model \wrt. the $i$-th token at the $j$-th layer (\ie, $h_{j,i}^{tea}$) :
\begin{equation}
    s_{l,i}(j) = -\left\Vert h_{l,i}^{stu} - W_j h_{j,i}^{tea}\right\Vert_2,
    \label{eq:similarity}
\end{equation}
where $j\in \mathcal{A}$, $\mathcal{A}$ represents the set of the teacher's layer indexes, and $W_j$ are learnable parameters.

Let $\mathcal{S}_{l,i}^K=\{s_{l,i}^k(\cdot)\}_{k=1}^K$ denote the top-$K$ similarities computed by Eq. (\ref{eq:similarity}) \wrt. $h_{l,i}^{stu}$.
We then train the student model by maximizing the similarities in $\mathcal{S}_{l,i}^K$.
Let $\beta_k$ denote the to-be-learned weighting scalar corresponding to the $k$-th similarity in $\mathcal{S}_{l,i}^K$.
The learning loss at the $l$-th layer \wrt. $\bm{x}$ can be formulated as:
\begin{equation}
    \label{equ:distill}
    \mathcal{L}_{(l)}^{\bm{x}} (\theta_{stu})=\frac{1}{N}\frac{1}{K} \sum_{i=1}^N  \sum_{k=1}^K - \beta_{k} \cdot s_{l,i}^k(\cdot).
\end{equation}

Finally, we integrate the prediction layer distillation and the intermediate layer distillation.
Let $\mathcal{T}$ denote the set of the student's layer indexes, 
the whole training loss of the student model is the summation of losses \wrt. all sentences in $\mathcal{D}_{in}$:
\begin{equation}
\small
    \mathcal{L} (\theta_{stu}) = \sum_{\bm{x}\in\mathcal{D}_{in}} \left(\lambda    \mathcal{L}_{pred}^{\bm{x}} + (1-\lambda)\sum_{l\in\mathcal{T}}\mathcal{L}_{(l)}^{\bm{x}}\right),
    \label{eq:loss}
\end{equation}
where $\lambda$ is a hyper-parameter for weighting. 

\subsubsection{Inference} For inference, we only use the learned student model $\theta_{stu}$ to compute perplexity for each token $x_i$ in an input sequence $\bm{x}=\{x_i\}_{i=1}^N$.
We calculate the OoD score \wrt. $\bm{x}$ by averaging over all tokens:
\begin{equation}
    \mathrm{score}(\bm{x}) = -\frac{1}{N}\sum_{i=1}^N \log p(x_i|x_{<i};\theta_{stu}).
\end{equation}

We define a threshold $\gamma$ to classify OoD examples against ID examples.
Specifically, $\bm{x}$ is predicted as an OoD example if $\mathrm{score}(\bm{x}) > \gamma$, else it is an ID example.

\begin{table*}[htb]
	\centering 
	\setlength{\tabcolsep}{0.6mm}
	\scalebox{0.81}{
	\begin{tabular}{l | c c c| c c c | ccc}  
		\toprule
        \multirow{2}{*}{Method}  & \multicolumn{3}{c|}{\textbf{CLINC150}} & \multicolumn{3}{c|}{\textbf{SST}} & \multicolumn{3}{c}{\textbf{ROSTD}}\\
        \cmidrule{2-4}\cmidrule{5-7}\cmidrule{8-10}
		& AUROC ($\uparrow$)  & AUPR ($\uparrow$)& FAR95 ($\downarrow$) & AUROC ($\uparrow$)  & AUPR ($\uparrow$)& FAR95 ($\downarrow$) & AUROC ($\uparrow$)  & AUPR ($\uparrow$)& FAR95 ($\downarrow$) \\
        \midrule
        TF-IDF+SVD$^{\dagger}$ & 58.5 & 21.8 & - & 78.0 & 73.2 & - & - & - & -\\
        Likelihood Ratios$^{\ddagger}$ & - & - & - & - & - & - & 96.35\small{\textpm 0.41} & 93.44\small{\textpm0.37} & 20.10\small{\textpm5.25} \\
        MDF+IMLM$^{\dagger}$ & 77.8 & 39.1 & - & 93.6 & 89.4 & -& - & - & -\\
        MDF+IMLM & 77.46\small{\textpm0.33} & 39.23\small{\textpm0.52} & 65.87\small{\textpm1.13} & 96.79\small{\textpm0.06} & 95.62\small{\textpm0.06} & 11.68\small{\textpm0.41} & 97.71\small{\textpm0.10} & 93.00\small{\textpm0.32} & 9.03\small{\textpm0.43} \\
        DATE & 83.38\small{\textpm0.15} & 50.21\small{\textpm0.18} & 66.67\small{\textpm1.65} & 82.20\small{\textpm0.18} & 83.11\small{\textpm0.41} & 55.26\small{\textpm1.97} & 96.59\small{\textpm0.43} & 91.77\small{\textpm1.06} & 17.06\small{\textpm1.82} \\
         \midrule
         M$_{\mathrm{finetune}}$  & 89.76\small{\textpm0.13} & 62.39\small{\textpm0.29} & 33.77\small{\textpm0.91} & 92.67\small{\textpm0.19} & 91.93\small{\textpm0.17} & 33.67\small{\textpm1.21} & 98.67\small{\textpm0.04} & 97.47\small{\textpm0.09} & 6.27\small{\textpm0.30}\\
         M$_{\mathrm{fromScratch}}$  & 91.73\small{\textpm0.12} & 68.78\small{\textpm0.62} & 28.31\small{\textpm0.40} & 96.60\small{\textpm0.65} & 96.42\small{\textpm0.63} & 17.98\small{\textpm3.47} & 99.10\small{\textpm0.03} & 98.25\small{\textpm0.06} & 3.88\small{\textpm0.22} \\
         Ours & \textbf{92.51\small{\textpm0.18}} & \textbf{70.94\small{\textpm0.78}} & \textbf{27.16\small{\textpm0.65}} & \textbf{97.97\small{\textpm0.40}} & \textbf{97.81\small{\textpm0.42}} & \textbf{9.50\small{\textpm2.09}} & \textbf{99.14\small{\textpm0.03}} & \textbf{98.33\small{\textpm0.06}} & \textbf{3.79\small{\textpm0.11}}\\
		\bottomrule
	\end{tabular}
    }
	\caption{Performance comparison on CLINC150, SST, and ROSTD. $^{\dagger}$ and $^{\ddagger}$ represents results reported in \citeauthor{xu2021unsupervised} \shortcite{xu2021unsupervised} and \citeauthor{Gangal2020LikelihoodRA} \shortcite{Gangal2020LikelihoodRA}, respectively.}
    \label{tab:clinc150_sst_rostd}
\end{table*}

\begin{table*}[htb]
	\centering 
	\setlength{\tabcolsep}{0.7mm}
	\scalebox{0.81}{
	\begin{tabular}{l | c c c| c c c | c c c}  
		\toprule
        \multirow{2}{*}{Method}  & \multicolumn{3}{c|}{\textbf{comp}} & \multicolumn{3}{c|}{\textbf{rec}} & \multicolumn{3}{c}{\textbf{sci}} \\
        \cmidrule{2-4}\cmidrule{5-7}\cmidrule{8-10}

        & AUROC ($\uparrow$)  & AUPR ($\uparrow$) & FAR95 ($\downarrow$) & AUROC ($\uparrow$)  & AUPR ($\uparrow$) & FAR95 ($\downarrow$) & AUROC ($\uparrow$)  & AUPR ($\uparrow$) & FAR95 ($\downarrow$)\\
        \midrule
        IsoForest$^{\dagger}$ & 66.1 & - & - & 59.4 & - & - & 57.8 & - & - \\
        OCSVM$^{\dagger}$ & 78.0 & - & - & 70.0 & - & - & 64.2 & - & - \\
        CVDD$^{\dagger}$ & 74.0 & - & - & 60.6 & - & - & 58.2 & - & - \\
        DATE$^{\dagger}$ & 92.1 & - & - & 83.4 & - & - & 69.7 & - & - \\
        DATE & 92.04\small{\textpm0.14} & 97.05\small{\textpm0.38} & 46.56\small{\textpm1.37} & 83.09\small{\textpm0.22} & 95.46\small{\textpm0.74} & 65.62\small{\textpm2.17} 
        & 66.30\small{\textpm0.16} & 90.64\small{\textpm0.12} & \textbf{81.68\small{\textpm0.67}} \\
        MDF + IMLM & 86.37\small{\textpm0.12} & 94.08\small{\textpm0.08} & 53.33\small{\textpm0.36} & 75.77\small{\textpm0.25} & 90.63\small{\textpm0.13} & 69.63\small{\textpm0.25} 
        & 67.02\small{\textpm0.12} & 87.81\small{\textpm0.10} & 86.18\small{\textpm0.68} \\
         \midrule
         M$_{\mathrm{finetune}}$  & 87.60\small{\textpm0.22} & 95.14\small{\textpm0.09} & 54.98\small{\textpm0.88} & 74.41\small{\textpm0.25} & 92.45\small{\textpm0.06} & 73.04\small{\textpm0.31}
         & 63.15\small{\textpm0.24} & 88.89\small{\textpm0.11} & 86.21\small{\textpm0.34} \\
         M$_{\mathrm{fromScratch}}$  & 91.29\small{\textpm0.60} & 97.29\small{\textpm0.11} & 49.36\small{\textpm5.08} & 85.36\small{\textpm0.56} & 96.27\small{\textpm0.12} & 64.28\small{\textpm3.66} 
         & 67.80\small{\textpm0.23} & 90.40\small{\textpm0.16} & 85.12\small{\textpm0.23} \\
         Ours &  \textbf{92.41\small{\textpm0.22}} & \textbf{97.49\small{\textpm0.12}} & \textbf{43.30\small{\textpm1.55}} & \textbf{87.68\small{\textpm0.15}} & \textbf{96.79\small{\textpm0.08}} & \textbf{54.66\small{\textpm1.41}}
         & \textbf{69.83\small{\textpm0.29}} & \textbf{91.14\small{\textpm0.15}} & 84.95\small{\textpm0.54}\\
         \midrule
         \midrule
        \multirow{2}{*}{Method}  & \multicolumn{3}{c|}{\textbf{misc}} & \multicolumn{3}{c}{\textbf{pol}} & \multicolumn{3}{c}{\textbf{rel}}\\
        \cmidrule{2-4}\cmidrule{5-7}\cmidrule{8-10}
         & AUROC ($\uparrow$)  & AUPR ($\uparrow$)& FAR95 ($\downarrow$) & AUROC ($\uparrow$)  & AUPR ($\uparrow$)& FAR95 ($\downarrow$) & AUROC ($\uparrow$)  & AUPR ($\uparrow$)&FAR95 ($\downarrow$) \\
        \midrule
        IsoForest$^{\dagger}$ & 62.4 & - & - & 65.3 & - & - & 71.4 & - & -\\
        OCSVM$^{\dagger}$ & 62.1 & - & - & 76.1 & - & - & 78.9 & - & -\\
        CVDD$^{\dagger}$ & 75.7 & - & - & 71.5 & - & - & 78.1 & - & -\\
        DATE$^{\dagger}$ & 86.0 & - & -  & 81.9 & - & - & 86.1 & - & -\\
        DATE & 82.25\small{\textpm0.12} & 98.93\small{\textpm0.01} & 66.81\small{\textpm2.40} & 81.74\small{\textpm0.16} & 96.72\small{\textpm0.10} & 64.52\small{\textpm2.72} & 86.14\small{\textpm0.09} & 97.66\small{\textpm0.02} & 62.86\small{\textpm0.93}\\
        MDF + IMLM & 62.26\small{\textpm7.70} & 96.80\small{\textpm0.81} & 92.81\small{\textpm5.10} & 81.05\small{\textpm0.16} & 95.48\small{\textpm0.06} & 62.38\small{\textpm0.31} & 80.85\small{\textpm0.26} & 96.14\small{\textpm0.08} & 65.14\small{\textpm0.35}\\
         \midrule
         M$_{\mathrm{finetune}}$ & 83.27\small{\textpm0.11} & 98.93\small{\textpm0.02} & 56.28\small{\textpm0.72} & 75.37\small{\textpm0.33} & 95.42\small{\textpm0.09} & 71.32\small{\textpm0.44} & 75.75\small{\textpm0.27} & 95.82\small{\textpm0.07} & 75.53\small{\textpm0.49}\\
         M$_{\mathrm{fromScratch}}$ & 85.01\small{\textpm0.25} & 99.16\small{\textpm0.01} & 63.56\small{\textpm2.00} & 87.53\small{\textpm0.17} & 97.87\small{\textpm0.01} & 51.32\small{\textpm0.62} & 86.48\small{\textpm0.18} & 97.90\small{\textpm0.06} & 58.57\small{\textpm1.23}\\
         Ours & \textbf{89.02\small{\textpm0.24}} & \textbf{99.35\small{\textpm0.03}} & \textbf{44.66\small{\textpm0.73}} & \textbf{88.47\small{\textpm0.15}} & \textbf{98.11\small{\textpm0.04}} & \textbf{50.85\small{\textpm1.07}} & \textbf{87.51\small{\textpm0.11}} & \textbf{98.01\small{\textpm0.03}} & \textbf{55.74\small{\textpm0.92}}\\
		\bottomrule
	\end{tabular}
    }
	\caption{Performance comparison on 20NewsGroups. $^{\dagger}$ represents results reported in \citeauthor{manolache2021date} \shortcite{manolache2021date}.}
    \label{tab:20newsgroups}
\end{table*}


\section{Experiments}

\subsection{Settings}
\label{sec:settings}
\paragraph{Datasets}
Following \citeauthor{xu2021unsupervised} \shortcite{xu2021unsupervised}, \citeauthor{manolache2021date} \shortcite{manolache2021date}, and \citeauthor{Gangal2020LikelihoodRA} \shortcite{Gangal2020LikelihoodRA}, we conduct five groups of experiments for evaluation: \textbf{CLINC150} \cite{larson2019evaluation},
\textbf{SST} \cite{socher2013recursive},
\textbf{ROSTD} \cite{Gangal2020LikelihoodRA},
\textbf{20NewsGroups} \cite{lang1995newsweeder}, and
\textbf{AGNews} \cite{zhang2015character}.
For dataset statistics and other detailed information, please refer to Appendix \ref{sec:appendix_dataset}.

\paragraph{Evaluation}
Following \citeauthor{hendrycks2017a} \shortcite{hendrycks2017a} and \citeauthor{jin2022towards} \shortcite{jin2022towards}, 
we utilize the Area Under Operating Characteristic curve (AUROC), the Area Under Precision-Recall curve (AUPR), and the False Alarm Rate (\ie, False Positive Rate) at 95\% Recall (FAR95) as metrics for a comprehensive evaluation.
Since our target is to detect OoD examples without having to rely on 
the semantic category of ID data, we treat OoD as the positive class for computing AUROC, AUPR, and FAR95.

\paragraph{Implementation Details}
We implement our approach based on PyTorch 1.12.0\footnote{\url{https://pytorch.org/}} and HuggingFace's Transformers\footnote{\url{https://github.com/huggingface/transformers}}.
For fair comparison, we utilize \texttt{GPT2-small} \cite{radford2019language} as the base model as it has a similar number of parameters as BERT-base \cite{devlin2019bert} used in \citeauthor{xu2021unsupervised} \shortcite{xu2021unsupervised} and the discriminator of ELECTRA \cite{clark2020ELECTRA} used in \citeauthor{manolache2021date} \shortcite{manolache2021date}.
Following \citeauthor{xu2021unsupervised} \shortcite{xu2021unsupervised}, we train both the teacher model and the student model for $5$ epochs on CLINC150, SST, and ROSTD.
Following  \citeauthor{manolache2021date} \shortcite{manolache2021date}, we train our models for $30$ epochs on 20 NewsGroups and $5$ epochs on AG News, respectively.
We use a batch size of $8$ and a maximum sequence length of $128$.
For the optimizers, we use AdamW~\citep{loshchilov2018decoupled} with the learning rate set to $5e-5$ for all models.
$\lambda$ in Eq. (\ref{eq:loss}) is set to $0.5$.
Motivated by \citeauthor{haidar-etal-2022-rail} \shortcite{haidar-etal-2022-rail}, which indicate that using only intermediate layers for distillation (from RoBERTa\_24 to RoBERTa\_6) works the best, we only distill the intermediate layers ($\mathcal{T}=\{l\}_{l=3}^9$).
We compare each student layer to a combination of K teacher layers, as  \citeauthor{haidar-etal-2022-rail} \shortcite{haidar-etal-2022-rail} show that concatenated representation distillation of sorted randomly selected K intermediate layers is superior to layer-wise distillation.
We choose $K=2$ for the cardinality of the similar set $\mathcal{S}_{l, i}^K$ considering that there's no information fusion among different teacher layers if $K=1$ and that a larger K may introduce too much noise due to the weighted average of representations as in Equation (\ref{equ:distill}).
We re-implement \citeauthor{xu2021unsupervised} \shortcite{xu2021unsupervised} and \citeauthor{manolache2021date} \shortcite{manolache2021date} with BERT-base using their open-sourced code, and report the results on all benchmark datasets for a more comprehensive comparison.
All experiments are conducted on one Tesla V100 (16GB). The trainable parameters (\ie, $\theta_{tea}$ and $\theta_{stu}$) are 248M. The training time is about 30 minutes for each model.

\begin{figure*}[htb]
  \centering
  \subfloat[Pre-trained GPT-2]{
    \label{sfig:gpt_embed}
    \includegraphics[height=3.2cm]{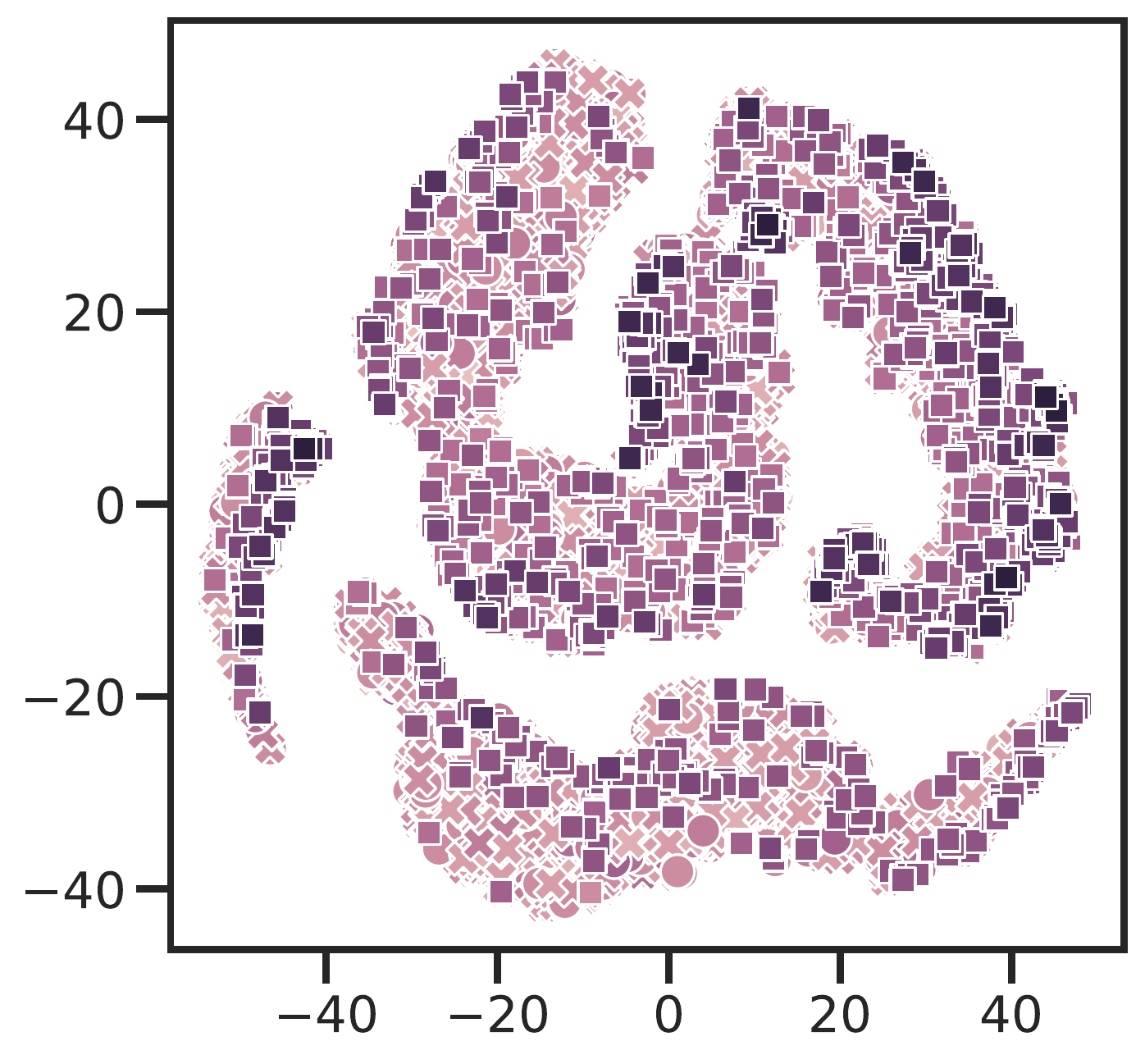}}
  \subfloat[M$_{\mathrm{finetune}}$]{
    \label{sfig:clm_ft_embed}
    \includegraphics[height=3.2cm]{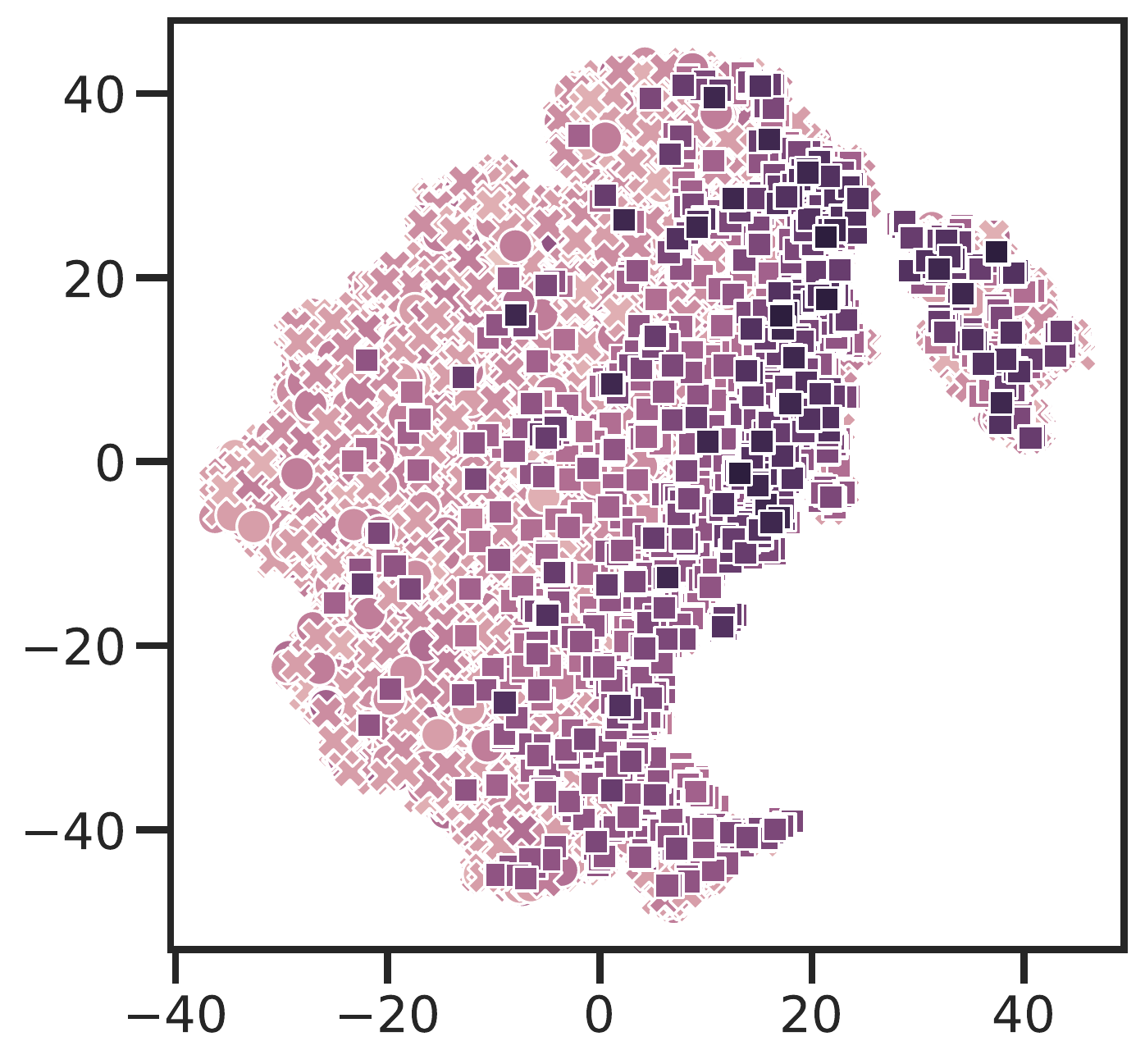}}
  \subfloat[M$_{\mathrm{fromScrath}}$]{
    \label{sfig:clm_scratch_embed}
    \includegraphics[height=3.2cm]{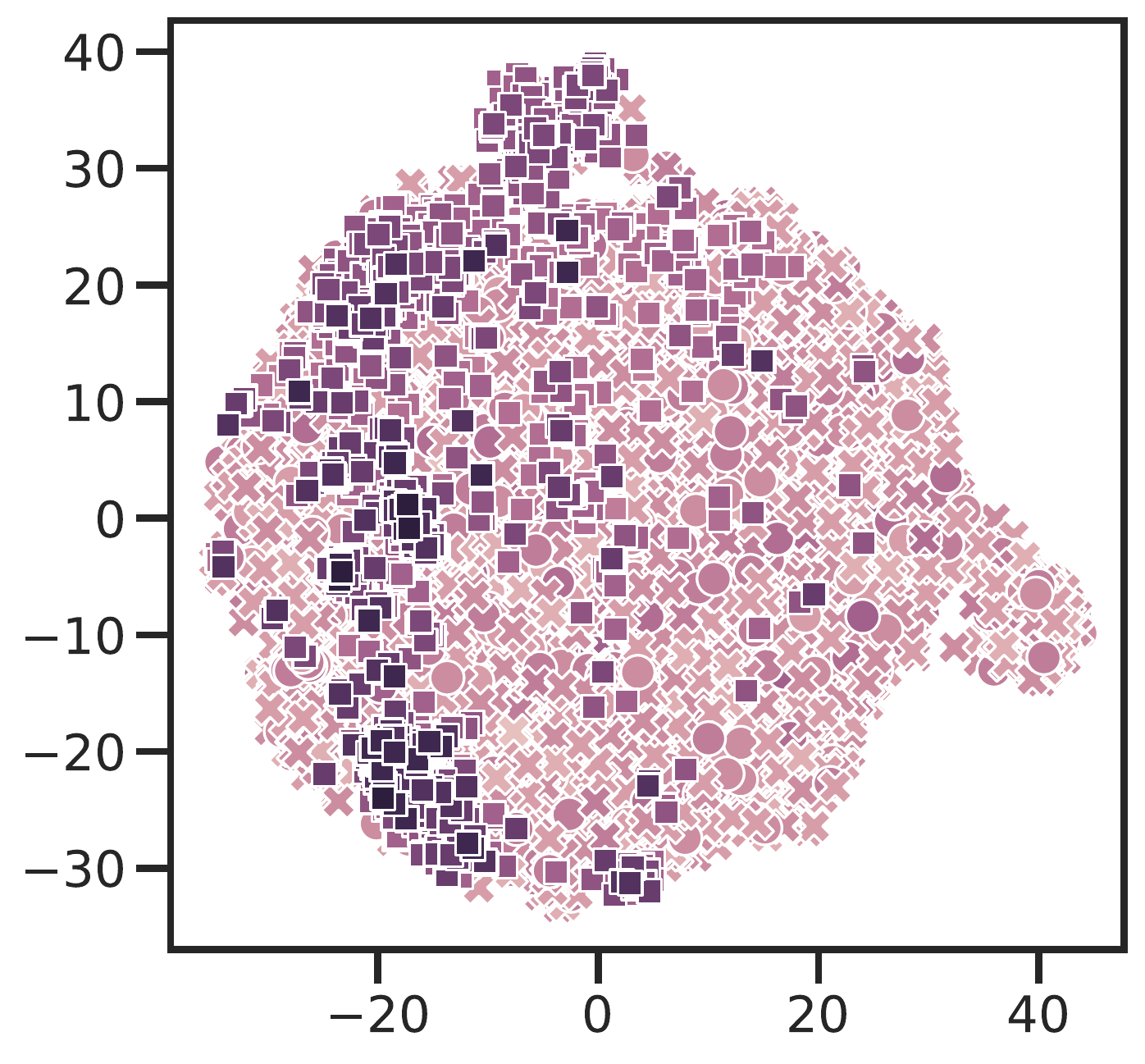}}
  \subfloat[Ours]{
    \label{sfig:ours_embed}
    \includegraphics[height=3.2cm]{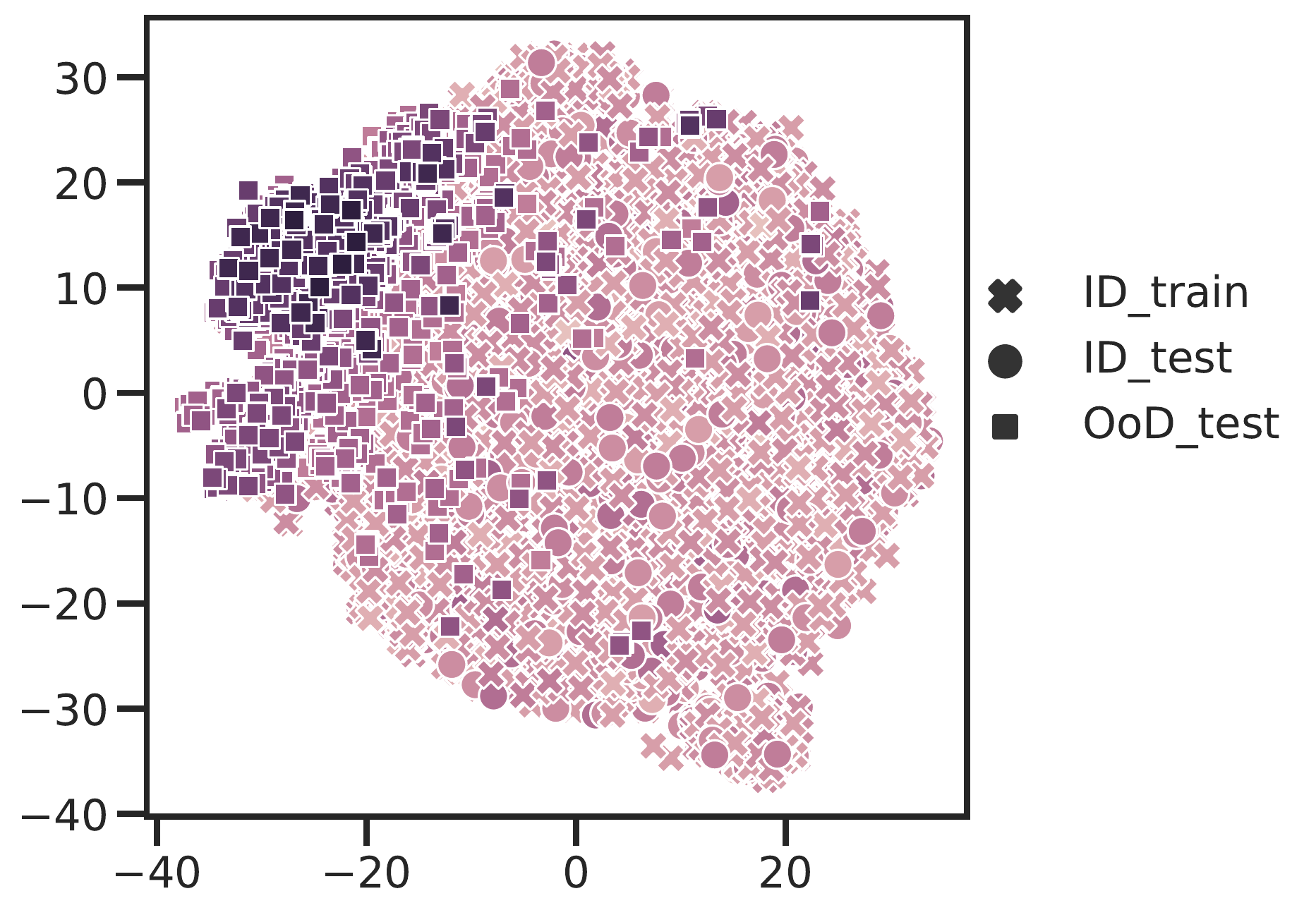}}
  \caption{T-SNE Visualization of sentence representations from different models. Darker colors represents higher OoD scores. }
  \label{fig:embed}
\end{figure*}

\subsection{Main Results}
Tables \ref{tab:clinc150_sst_rostd} and \ref{tab:20newsgroups} report the results of our approach alongside those reported by previous state-of-the-art methods on CLINC150, SST, ROSTD, and 20NewsGroups.
It can be seen that our proposed method outperforms the prior methods with a large margin in most experiments, achieving an improvement of up to 9.13, 20.73, 38.71 points in terms of AUROC, AUPR, and FAR95, respectively, on CLINC150.
This well demonstrates the effectiveness of the proposed approach.

The results also show that M$_{\mathrm{fromScrath}}$ generally leads to superior performance than M$_{\mathrm{finetune}}$.
We conjecture that seeing no OoD examples during parameter learning helps the randomly initialized model avoid optimizing toward OoD distribution.
Without the bias and constraints inherited from the pre-training process, the model trained from scratch is more likely to find a local minimum that better fits the ID training text and thus leads to more distinguishable features for OoD detection.
Moreover, our approach, which uses the fine-tuned model to teach a randomly initialized model, can integrate their strengths via the proposed multi-level knowledge distillation process, resulting in superior performance.


\subsection{Ablation Study}
To validate the contributions of different components in the proposed approach, here we introduce two variants of our model 
for ablation study:
i) \textit{Ours w/ GPT2\_Init\_$\theta_{stu}$}, which initializes the student model with the pre-trained GPT-2 model.
ii) \textit{Ours w/o $\mathcal{L}^{\bm{x}}_{(l)}$}, which eliminates the loss \wrt. intermediate layer distillation and only conducts output layer distillation to learn the student model.
Table \ref{tab:ablation_inter} shows the results.
\begin{table}[h]
    \centering
	\setlength{\tabcolsep}{0.5mm}
	\scalebox{0.8}{
    \begin{tabular}{lc c c}
    \toprule
         & AUROC ($\uparrow$) & AUPR ($\uparrow$) & FAR95 ($\downarrow$) \\
         \midrule
        Ours & \textbf{97.97\small{\textpm0.40}} & \textbf{97.81\small{\textpm0.42}} & \textbf{9.50\small{\textpm2.09}}\\
        Ours w/ GPT2\_Init\_$\theta_{stu}$ & 94.12\small{\textpm0.60} & 94.21\small{\textpm0.64} & 31.72\small{\textpm3.20} \\
        Ours w/o $\mathcal{L}_{(l)}^{\bm{x}}$  & 97.07\small{\textpm0.23} & 96.94\small{\textpm0.23} & 14.53\small{\textpm1.05} \\
        \bottomrule
    \end{tabular}
    }
    \caption{Ablation study on SST.}
    \label{tab:ablation_inter}
\end{table}

Comparing Ours with \textit{Ours w/ GPT2\_Init\_$\theta_{stu}$}, we can see that replacing the randomly initialized student model with a pre-trained student model will cause a significant performance drop, well verifying our motivation to incorporate M$_{\mathrm{fromScratch}}$ with M$_{\mathrm{finetune}}$.
Table \ref{tab:ablation_inter} also illustrates that removing the constraints on intermediate layers, \ie, $\mathcal{L}_{(l)}^{(\bm{x})}$, the student model's performance will decrease by 0.90, 0.87, and 5.03 in terms of AUROC, AUPR, and FAR95, respectively.
This well validates both the effectiveness and necessity of intermediate layer distillation.
Moreover, though eliminating the intermediate distillation, the student model in \textit{Ours - $\mathcal{L}_{(l)}^{\bm{x}}$} which is derived with only the prediction layer distillation still outperforms the baseline model M$_{\mathrm{fromScratch}}$.
We owe this superiority to the more informative supervision, \ie, the probability distribution produced by the teacher model, compared with the ground-truth one-hot supervision used in M$_{\mathrm{fromScratch}}$.

\subsection{Analysis on Distribution of Sentence Repr.}
To bring up insights on how multi-level knowledge distillation promotes OoD detection, we utilize \textit{t-SNE} \cite{van2008visualizing} to reduce the dimension of sentence representations obtained from pre-trained GPT-2, M$_{\mathrm{finetune}}$, M$_{\mathrm{fromScratch}}$, and the student model in our approach.
Here, we produce sentence representations by averaging token representations.
The visualization is shown in Figure \ref{fig:embed}.
In Figure (\ref{sfig:gpt_embed}), we can see that ID and OoD examples locate uniformly in several \textit{separate} data manifolds because the model has no sense of what is OoD and thus struggles to distinguish OoD examples from ID examples.
After applying fine-tuning to the pre-trained model, representations of the ID data converge to fewer manifolds and it becomes easier to classify ID and OoD examples, specifically, OoD examples mainly lie in the right-side of Figure (\ref{sfig:clm_ft_embed}), which might lead to easier separation in high dimension space.
However, when comparing Figures (\ref{sfig:clm_ft_embed}) and (\ref{sfig:clm_scratch_embed}), we can notice that M$_{\mathrm{finetune}}$ doesn't fit the ID examples as well as M$_{\mathrm{fromScratch}}$ because the representation distribution of ID examples in Figure (\ref{sfig:clm_scratch_embed}) is more condensed.
This supports our conjecture that the pre-training process may guide the model to a position near an inferior local minimum where ID and OoD examples are less separable.
Last but not least, Figure (\ref{sfig:ours_embed}) indicates that our student model produces a more compact distribution for ID examples when trained from scratch.
Meanwhile, Figure (\ref{sfig:ours_embed}) also shows that ID representations and OoD representations produced by our student model are more dissociable.
We conjecture that this is because the model gains some ability to separate examples of different semantics via knowledge distillation - the teacher model equips this knowledge during pre-training.

\subsection{Application}
\label{sec:application}
ChatGPT, an optimized language model for dialog\footnote{\url{https://chat.openai.com/}}, has attracted great attention in the NLP field since its inception.
It is capable of providing fluent and comprehensive responses to a large variety of questions.
To study how close ChatGPT is to human experts, \citet{guo2023close} proposed the Human ChatGPT Comparison Corpus (HC3), where each question is paired with two answers, one is a human answer collected from wiki sources and public QA datasets, and the other is generated by ChatGPT\footnote{HC3 covers a wide range of domains (open-domain, computer science, finance, medicine, law, and psychology) and is widely used in lots of recent studies.}.
By conducting human evaluation, \citet{guo2023close} indicates that it can be difficult to distinguish texts generated by ChatGPT from those provided by human experts, and further propose a RoBERTa \cite{Liu2019RoBERTaAR} based detector to distinguish both.

Following \citet{guo2023close}, in this section, we adapt our model as an AI-generated content (AIGC) detector to explore its capability for preventing the potential risks of AIGC abuse.
As our model uses perplexity as the OoD score and \citet{guo2023close} reveal that ChatGPT-generated answers are usually of low perplexities, here we take ChatGPT-generated answers as in-distribution data to train our model.
We divide the in-distribution data into a training set and a test set.
We  use all the human-generated answers as the OoD test set.

We first evaluate our model as in \ref{sec:settings} and Table \ref{tab:hc3} shows its performance results.
We can see that our approach significantly outperforms prior state-of-the-art methods \textit{DATE} and \textit{MDF+IMLM} under the same settings.
Surprisingly, our unsupervised method demonstrates comparable performance with \textit{RoBERTa-single Detector}, which is a RoBERTa-based sentence classifier trained with the supervision from all the ChatGPT-generated and human-generated texts.

\begin{table}[htb]
	\centering 
	\setlength{\tabcolsep}{0.5mm}
	\scalebox{0.7}{
	\begin{tabular}{l  c c c}  
	\toprule
          &  AUROC ($\uparrow$)  & AUPR ($\uparrow$) & FAR95 ($\downarrow$)  \\
        \midrule
        \textit{Unsupervised methods:} & & &\\
        DATE & 75.80 & 91.20 & 85.15  \\
        MDF+IMLM (BERT) & 89.61 & 96.80 & 42.35  \\
        MDF+IMLM (GPT2-small) & 91.53 & 92.56 & 31.84  \\
                 \textbf{Ours} &  \textbf{99.80} & \textbf{99.95} & \textbf{0.61} \\
        \midrule
        \textit{Supervised method:} & & &\\
        chatgpt-detector-roberta\footnotemark & 99.98 & 99.99 & 0.04 \\
	\bottomrule
	\end{tabular}
    }
	\caption{Performance comparison on HC3.}
    \label{tab:hc3}
\end{table}

\footnotetext{\url{https://github.com/Hello-SimpleAI/chatgpt-comparison-detection}}

We also compare our model to the human evaluation results listed in \citet{guo2023close}.
Given two answers corresponding to the same question, with one being generated by ChatGPT and the other by a human expert,
our model is required to determine which answer is generated by ChatGPT.
Table \ref{tab:human} shows that our model beats human evaluators and perfectly handles this task.
\begin{table}[htb]
	\centering 
	\setlength{\tabcolsep}{5mm}
	\scalebox{0.8}{
	\begin{tabular}{l  c c}  
	\toprule
         &  Human & Ours \\
        \midrule
        \textbf{All} & 0.90 & \textbf{1.00} \\
        \midrule
        reddit\_eli5 & 0.97 & -  \\
        open\_qa & 0.98 & - \\
        wiki\_csai & 0.97 & - \\
        medical & 0.97 & - \\
        finance & 0.79 & - \\
	\bottomrule
	\end{tabular}
    }
	\caption{Accuracy comparison with human evaluation on paired answers to determine the ChatGPT-generated responses. Note that we run experiments using the whole test set, while human evaluation in \citet{guo2023close} is conducted on a subset of it.}
    \label{tab:human}
\end{table}


\begin{table*}[t]
	\centering 
	\setlength{\tabcolsep}{1.5mm}
		\scalebox{0.85}{

 	\begin{tabular}{l | c c c| c c c}
		\toprule
        \multirow{2}{*}{Method}  & \multicolumn{3}{c|}{\textbf{business}} & \multicolumn{3}{c}{\textbf{sci}}\\
        \cmidrule{2-4}\cmidrule{5-7}
		& AUROC ($\uparrow$)  & AUPR ($\uparrow$)& FAR95 ($\downarrow$) & AUROC ($\uparrow$)  & AUPR ($\uparrow$)& FAR95 ($\downarrow$) \\
        \midrule
        IsoForest$^{\dagger}$ & 73.2& - & - & 76.9 & - & -\\
        OCSVM$^{\dagger}$ & 83.2 & - & - & 80.7 & - & -\\
        CVDD$^{\dagger}$ & 79.6 & - & - & 79.0 & - & -\\
        DATE$^{\dagger}$ & 90.1 & - & - & 84.0 & - & -\\
        DATE & 89.46\small{\textpm0.15} & 95.19\small{\textpm0.12} & 50.26\small{\textpm1.49} & 83.88\small{\textpm0.37} & 93.29\small{\textpm0.24} & 60.97\small{\textpm3.12}\\
        MDF + IMLM & 90.12\small{\textpm0.06} & 95.36\small{\textpm0.03} & 34.81\small{\textpm0.39} & \textbf{85.93\small{\textpm0.22}} & \textbf{94.62\small{\textpm0.09}} & 54.37\small{\textpm1.30}\\
         \midrule
         M$_{\mathrm{finetune}}$  & 89.19\small{\textpm0.10} & 95.27\small{\textpm0.05} & 67.63\small{\textpm0.24} & 76.52\small{\textpm0.14} & 88.36\small{\textpm0.13} & 63.95\small{\textpm0.42}\\
         M$_{\mathrm{fromScratch}}$  & 91.49\small{\textpm0.16} & 96.29\small{\textpm0.08} & 32.56\small{\textpm0.87} & 83.76\small{\textpm0.09} & 92.12\small{\textpm0.07} & 54.42\small{\textpm0.31}\\
         Ours & \textbf{92.38\small{\textpm0.11}} & \textbf{96.72\small{\textpm0.05}} & \textbf{29.94\small{\textpm0.68}} & 85.12\small{\textpm0.19} & 92.92\small{\textpm0.16} & \textbf{51.18\small{\textpm0.38}}\\
         \midrule
         \midrule
        \multirow{2}{*}{Method}  & \multicolumn{3}{c|}{\textbf{sports}} & \multicolumn{3}{c}{\textbf{world}}\\
        \cmidrule{2-4}\cmidrule{5-7}
		& AUROC ($\uparrow$)  & AUPR ($\uparrow$)& FAR95 ($\downarrow$) & AUROC ($\uparrow$)  & AUPR ($\uparrow$)& FAR95 ($\downarrow$) \\
        \midrule
        IsoForest$^{\dagger}$ & 84.7 & - & - & 79.6 & - & -\\
        OCSVM$^{\dagger}$ & 92.4 & - & - & 79.9 & - & -\\
        CVDD$^{\dagger}$ & 89.9 & - & - & 84.0 & - & -\\
        DATE$^{\dagger}$ & 95.9 & - & & 90.0 & - & -\\
        DATE & 96.01\small{\textpm0.22} & 98.48\small{\textpm0.08} & 19.21\small{\textpm0.68} & 90.08\small{\textpm0.14} & 96.04\small{\textpm0.06} & 42.46\small{\textpm0.38}\\
        MDF + IMLM & \textbf{97.91\small{\textpm0.06}} & \textbf{99.20\small{\textpm0.03}} & \textbf{7.37\small{\textpm0.23}} & \textbf{91.28\small{\textpm0.08}} & \textbf{96.68\small{\textpm0.04}} & 38.63\small{\textpm0.47}\\
         \midrule
         M$_{\mathrm{finetune}}$  & 91.46\small{\textpm0.19} & 96.09\small{\textpm0.11} & 30.79\small{\textpm0.52} & 84.19\small{\textpm0.07} & 92.17\small{\textpm0.05} & 48.24\small{\textpm0.42}\\
         M$_{\mathrm{fromScratch}}$  & 97.00\small{\textpm0.08} & 98.86\small{\textpm0.04} & 10.75\small{\textpm0.39} & 89.46\small{\textpm0.05} & 95.27\small{\textpm0.04} & 38.51\small{\textpm0.59}\\
         Ours & 97.31\small{\textpm0.06} & 98.97\small{\textpm0.03} & 9.27\small{\textpm0.36} & 89.89\small{\textpm0.04} & 95.43\small{\textpm0.04} & \textbf{37.86\small{\textpm0.35}}\\
		\bottomrule
	\end{tabular}
    }
	\caption{Performance comparison on AGNews. $^{\dagger}$ represents results reported in \citeauthor{manolache2021date} \shortcite{manolache2021date}.}
    \label{tab:agnews}
\end{table*}

\section{Conclusion}
In this paper, we focus on the setting of OoD detection without supervision from both OoD data nor ID class labels.
We analyze the complementary characteristics of existing self-supervised representation learning-based methods and propose a multi-level knowledge distillation approach to integrate their strengths, while mitigating their limitations.
We evaluate the proposed method on multiple datasets and results show that the proposed method yields new state-of-the-art performance.
We analyze why our approach attains superior performance by conducting ablation studies and sentence representation visualization.
We further apply our model as an AIGC detector to distinguish ChatGPT-generated texts from those generated by human experts and the experimental results demonstrate that our model outperforms human evaluators in the setting of paired answers.

\section*{Limitations}
Table \ref{tab:agnews} shows the results of our model and other methods on the AGNews benchmark.
Interestingly, we notice that our approach reports a slightly inferior performance when compared with \textit{MDF+IMLM} \cite{xu2021unsupervised}.
We can see that methods using sentence representations based on token aggregation, \eg, fastText\footnote{\url{https://github.com/facebookresearch/fastText}} or Glove \cite{pennington2014glove}-based IsoForest, OCSVM, and CVDD \cite{ruff2019self}, as well as BERT based MDF + IMLM \cite{xu2021unsupervised}, perform especially well on AGNews compared to their performance on other datasets.
We conjecture that this is because AGNews has a much larger variation of sequence length (36.6) than other datasets (around 7 or 8).
A larger length variation will lead to more acute fluctuations in perplexities, especially when adopting an autoregressive language model with unidirectional context such as GPT-2-small in this paper, making it more difficult to distinguish between ID and OOD examples than in other datasets.
In contrast, sentence representation based methods benefit from directly estimating the OoD score using the information from the whole sentence, thus producing superior performance.
Fortunately, the limitation of auto-regressive modeling could be eliminated by leveraging Transcormer \cite{song2022transcormer} as the base model of our approach, where bidirectional context is used for estimating tokens at each position. We leave this for future work.

\bibliography{anthology,custom}
\bibliographystyle{acl_natbib}

\newpage
\appendix
\section{Appendix}

\subsection{Dataset Details}
\label{sec:appendix_dataset}
\begin{itemize}
    \item \textbf{Group 1: CLINC150.} \citeauthor{larson2019evaluation} \shortcite{larson2019evaluation} introduce a crowdsourced dialog dataset. Following \citeauthor{xu2021unsupervised} \shortcite{xu2021unsupervised}, we use all training queries covering 150 intents as ID training data and fuse 4500 ID examples of the test split with 1000 OoD examples for evaluation.
    \item \textbf{Group 2: SST.} Following \citeauthor{hendrycks2020pretrained} \shortcite{hendrycks2020pretrained} and \citeauthor{xu2021unsupervised} \shortcite{xu2021unsupervised}, we use the training split of the SST dataset \cite{socher2013recursive} for ID training examples and use its test split as ID test examples. The same described random sample of 500 examples from 20 NewsGroups \cite{lang1995newsweeder}, English-German Nulti30K \cite{elliott2016multi30k}, RTE \cite{dagan2005thePR}, and SNLI \cite{bowman2015large}, is combined and used as OoD test examples.
    \item \textbf{Group 3: ROSTD.} \citet{Gangal2020LikelihoodRA} releases a dataset consisting of 4590 OoD examples with respect to the English split of \citet{schuster2019cross} as the ID dataset. Here we use the training ID, test-ID, and actual OoD as in \citet{Gangal2020LikelihoodRA} for fair comparison.
    \item \textbf{Group 4: 20NewsGroups.} Following \citeauthor{manolache2021date} \shortcite{manolache2021date}, we only consider articles from six top-level classes of 20NewsGroups \cite{lang1995newsweeder} for evaluation. We construct the ID data using examples from a single label, \ie, training split for ID training and test split for ID test. We take data corresponding to other labels in the test split as OoD test examples.
    \item \textbf{Group 5: AGNews.} AG News is a topic classification dataset \cite{zhang2015character} collected from various news sources. There are four topics in total.
    Similar to Group 4, we conduct experiments with each single label for ID and others for OoD, respectively.
    \item \textbf{Group 6: HC3.} Human ChatGPT Comparison Corpus (HC3) is a question-answer dataset ~\cite{guo2023close}, which collects the human (from wiki and public QA datasets) and ChatGPT answers for the same questions.
    HC3 covers a wide range of domains (open-domain, computer science, finance, medicine, law, and psychology).
    We conduct experiments with ChatGPT answers for ID and human answers for OoD at the sentence level, respectively.
\end{itemize}

Table \ref{tab:data_statistics} shows the statistics of the different datasets and sub-topics, if any.
\begin{table}[ht!]
	\centering 
	\setlength{\tabcolsep}{1.5mm}
        \scalebox{0.8}{
	\begin{tabular}{c l | c c c}  
		\toprule
		\multicolumn{2}{c|}{\multirow{2}{*}{\textbf{Group}}} & \textbf{\# of ID} & \textbf{\# of ID} & \textbf{\# of OoD}\\
		& & (train) & (test)  & (test)\\
        \midrule
        \multicolumn{2}{c|}{\#1: CLINC150} & 15000 & 4500 & 1000 \\
        \midrule
        \multicolumn{2}{c|}{\#2: SST} & 8544 & 2210 & 2000 \\
        \midrule
        \multicolumn{2}{c|}{\#3: ROSTD} & 30521 & 8621 & 4590 \\
        \midrule
        & comp & 2857 & 1909 & 5390 \\
        & misc & 577 & 382 & 6917\\
        \#4: & pol & 1531 & 1025 & 6274\\
        20NewsGroups & rec & 2301 & 1524 & 5775\\
        & rel & 1419 & 939 & 6360\\
        & sci & 2311 & 1520 & 5779\\
        \midrule
        & business & 30000 & 1900 & 5700\\
        \#5: & sci & 30000 & 1900 & 5700\\
        AGNews & sports & 30000 & 1900 & 5700\\
        & world & 30000 & 1900 & 5700\\
        \midrule
        \multicolumn{2}{c|}{\#6: HC3} & 13442 & 13443 & 58546 \\
		\bottomrule
	\end{tabular}
        }	
 \caption{Dataset statistics.}
    \label{tab:data_statistics}
\end{table}

\subsection{\textit{MDF+IMLM} with Different Base Models} 
We take MDF+IMLM from \citeauthor{xu2021unsupervised} \shortcite{xu2021unsupervised} as one of the baselines.
In the main body of this paper, we show the results of MDF+IMLM with BERT as the base model because BERT is the most considered counterpart for GPT-2-small used in our approach.
Here we include the RoBERTa-based results of MDF+IMLM from \citeauthor{xu2021unsupervised} \shortcite{xu2021unsupervised} for your information.

\begin{table*}[htb]
	\centering 
	\scalebox{0.9}{
	\begin{tabular}{l | c c c| c c c}  
		\toprule
        \multirow{2}{*}{Method}  & \multicolumn{3}{c|}{\textbf{CLINC150}} & \multicolumn{3}{c}{\textbf{SST}}\\
        \cmidrule{2-4}\cmidrule{5-7}
		& AUROC ($\uparrow$)  & AUPR ($\uparrow$)& FAR95 ($\downarrow$) & AUROC ($\uparrow$)  & AUPR ($\uparrow$)& FAR95 ($\downarrow$) \\
        \midrule
        MDF+IMLM (GPT-2-small)$^{\ddagger}$ & 72.03 & 31.30 & 74.29 & - & - & -\\
        MDF+IMLM (BERT)$^{\dagger}$ & 77.8 & 39.1 & - & 93.6 & 89.4 & -\\
        MDF+IMLM (BERT)$^{\ddagger}$ & 77.46 & 39.23 & 65.87 & 96.79 & 95.62 & 11.68  \\
        MDF+IMLM (RoBERTa)$^{\dagger}$ & 80.1 & 44.9 & - & \textbf{99.9} & \textbf{99.8} & -\\
        \midrule
         Ours (GPT-2-small) & \textbf{92.51} & \textbf{70.94} & \textbf{27.16} & 97.97 & 97.81 & \textbf{9.50} \\
		\bottomrule
	\end{tabular}
    }
	\caption{Performance comparison on CLINC150 and SST. $^{\dagger}$ represents results reported in \citeauthor{xu2021unsupervised} \shortcite{xu2021unsupervised}. $^{\ddagger}$ denotes our re-implemented results.}
    \label{tab:mdf_base_model}
\end{table*}

Table \ref{tab:mdf_base_model} shows that using a more powerful base model does bring significant performance gain to MDF+IMLM.
Though our model is implemented with GPT-2-small, it still demonstrates comparable (on SST) and even superior performance (CLINC150) with RoBERTa based MDF+IMLM.

\subsection{Discussion on CLM and MLM.}
Here we discuss the consideration for using CLM rather than MLM.
In fact, we conducted experiments using the previous method of masking X\% of tokens for one forward.
However, the results were not satisfactory.
We attribute this to an insufficient perplexity estimation in a single forward.
In other words, with MLM, it would be better to recover the joint probability of the entire input sequence to achieve better performance, \ie, forwarding an input sentence multiple times so that the probability of each token in the sentence could be predicted.
This should be time-consuming and thus we use CLM in this paper.

\end{document}